\documentclass[runningheads]{llncs}
\usepackage[T1]{fontenc}
\usepackage{graphicx}
\usepackage{todonotes}
\usepackage{graphicx}
\usepackage{amsmath}
\usepackage{amssymb}
\usepackage{algorithm}
\usepackage{algpseudocode}
\usepackage{booktabs}
\usepackage{multirow}
\graphicspath{ {figures/} }
\usepackage{adjustbox}
\usepackage{caption}
\usepackage{subcaption}

\newcommand{\loss}{\mathcal{L}}

\DeclareMathOperator{\sg}{sg}

\begin{document}

\title{Blending Low and High-Level Semantics of Time Series for Better Masked Time Series Generation}
\titlerunning{Blending Low and High-Level Semantics of Time Series for Masked TSG}
%


\author{Johan Vik Mathisen$^{1,*}$ \and
Erlend Lokna$^{1,*}$ \and
Daesoo Lee\inst{1} \and
Erlend Aune\inst{1,2,3}}

\authorrunning{J. V. Mathisen, E. Lokna  et al.}
%

\institute{
Norwegian University of Science and Technology, Trondheim, Norway \\ \and
BI Norwegian Business School, Oslo, Norway \\ \and
HANCE, Oslo, Norway
}

\maketitle              

\def\thefootnote{*}\footnotetext{These authors contributed equally to this work.}\def\thefootnote{\arabic{footnote}}

\begin{abstract}

State-of-the-art approaches in time series generation (TSG), such as TimeVQVAE, utilize vector quantization-based tokenization to effectively model complex distributions of time series. 
These approaches first learn to transform time series into a sequence of discrete latent vectors, and then a prior model is learned to model the sequence.
The discrete latent vectors, however, only capture low-level semantics (\textit{e.g.,} shapes).
We hypothesize that higher-fidelity time series can be generated by training a prior model on more informative discrete latent vectors that contain both low and high-level semantics (\textit{e.g.,} characteristic dynamics).
In this paper, we introduce a novel framework, termed NC-VQVAE, to integrate self-supervised learning into those TSG methods to derive a discrete latent space where low and high-level semantics are captured.
Our experimental results demonstrate that NC-VQVAE results in a considerable improvement in the quality of synthetic samples.

\keywords{Time series  \and Self-supervised learning \and TimeVQVAE \and Masked modelling.}
\end{abstract}

\section{Introduction}

Time series generation (TSG) models have been created to address data limitations stemming from challenges in data acquisition or privacy restrictions. In recent years, the TSG field has grown its popularity \cite{lee2023vector,li2022tts,desai2021timevae}, exploring various approaches such as GAN \cite{goodfellow2020generative}, Variational AutoEncoder (VAE) \cite{Kingma2013AutoEncodingVB}, and Vector Quantized-Variational AutoEncoder (VQVAE) \cite{van2017neural}. The TSG survey paper \cite{ang2023tsgbench} demonstrated that the current state-of-the-art methods are VAE and VQ-VAE-based methods such as TimeVAE \cite{desai2021timevae} and TimeVQVAE \cite{lee2023vector}. 

Another important field is self-supervised learning (SSL) on time series. It has gained attention due to its efficiency in uncovering distinguishable patterns from data without requiring labeled datasets or human supervision. When trained with SSL, the model can produce informative representations (\textit{i.e.,} latent vectors) and they are often utilized to ease a downstream task such as classification, as demonstrated in \cite{lee2023ensemble}. 

While TSG and SSL have become two important branches of time series methodology, their efforts have been orthogonal. We hypothesize that TSG can benefit from quality representations learned with SSL, leading to higher fidelity of synthetic time series. This can be somewhat evidenced by several studies in computer vision and audio domains; for instance, \cite{RCG2023} demonstrated improved image generation capability by conditioning a representation during the sampling process, and \cite{borsos2023audiolm} showed a similar approach for audio data. 

Our study clearly differs from the existing approaches of employing SSL representations, which focus on representation conditioning. We examine the integration of TSG and SSL rather than utilizing the two as separate components. More precisely, we train a VQVAE model that employs a non-contrastive self-supervised loss-function in addition to the reconstruction and codebook losses. This enforces both low-level (e.g., local shapes) and high-level (e.g., characteristic dynamics) semantics into the latent space. Subsequently, we train a generative model on the learned latent space and examine the results along several dimensions.


On a subset of the UCR archive we demonstrate how our proposed model, NC-VQVAE, significantly improves the quality of the discrete latent representations, compared to naive VQVAE. Our results in terms of classification accuracy with SVM and KNN, inception score (IS), Fréchet inception distance (FID) score, and visual inspection, show that NC-VQVAE results in a considerable improvement in the quality of both latent representations and synthetic samples.




\section{Related Work}

Our work involves building a framework that leverage non-contrastive SSL to improve representation learning in a VQVAE \cite{VQVAE} type model. This section presents the works related to ours. To the best of our knowledge, enhancing VQ-based tokenization models with SSL methods has not yet been investigated in the time series domain. 
We base our model on a simplified version of TimeVQVAE \cite{lee2023vector}, which utilize a bidirectional transformer model, MaskGIT \cite{chang2022maskgit}, for prior learning. Although our non-contrastive SSL extension could use any model with a siamese architecture, we experiment with the proven methods Barlow Twins \cite{zbontar2021barlow} and VIbCReg \cite{lee2024computer}.

\subsection{Time Series Generation}

\subsubsection{VQVAE} \cite{VQVAE} consists of an encoder, decoder, and a codebook. The input data is mapped into a latent space by the encoder and then quantized through a nearest neighbor lookup in the codebook, producing discrete latent vectors. These vectors are subsequently mapped back by the decoder to reconstruct the original data. VQVAE employs a two-stage modeling approach: first, the encoder, decoder, and codebook are trained; then, all parameters are frozen, and a prior model is trained on the discrete latent space. This approach provides more flexibility in capturing the underlying data distribution while simultaneously avoiding posterior collapse and variance issues observed in VAEs.

\subsubsection{MaskGIT} \cite{chang2022maskgit} employs a bi-directional transformer for prior learning. During training, it uses masked modeling, where a random portion of the tokens are masked with a special mask token. MaskGIT learns to predict these masked tokens by attending to tokens in all directions, thereby learning the prior distribution of the latent space. During inference, MaskGIT starts with a blank canvas made entirely of mask tokens and iteratively predicts the entire image by conditioning on the most confident tokens from previous predictions.

\subsubsection{TimeVQVAE} \cite{lee2023vector} is a time series generation model based on VQVAE and MaskGIT. It employs a two-stage approach similar to VQVAE, using a bidirectional transformer for prior learning, akin to MaskGIT. The authors introduce vector quantization modeling in the time-frequency domain, separating data into high and low-frequency components to better retain temporal consistencies and generate higher-quality samples.

\subsection{Self-supervised Learning}

\subsubsection{Barlow Twins} \cite{zbontar2021barlow} is a non-contrastive SSL method. The key idea is to encourage representations of similar samples to be alike while reducing redundancy between the components of the vectors.
For each sample in a batch, two augmented views are first generated. Then, the model projects the batch of views onto a feature space such that the correlation between the views from the same sample is close to 1 and the correlation between the views of different samples is close to 0.
This approach enforces both distortion invariance and decorrelated features in the representations.

\subsubsection{VIbCReg} \cite{lee2024computer} is also a non-contrastive SSL model with a siamese architecture based on VICReg \cite{bardes2022vicreg}. It improves upon VICReg by incorporating better covariance regularization and IterNorm \cite{huang2019iterative}. A key difference from Barlow Twins is that variance and covariance regularizations are applied to each branch individually.
As with Barlow Twins, a batch is augmented to create two views and passed through an encoder and projector. The projector outputs are then \textit{whitened} using IterNorm before calculating its loss.

\section{Method}
The work in this article builds on \cite{lee2023vector}. We simplify the model architecture by omitting the high-low frequency split, reducing the model to what they refer to as naive VQVAE in their paper. We expand upon naive VQVAE by integrating a self-supervised learning loss in Stage 1 of the TimeVQVAE framework.

A schematic figure of our proposed vector quantization model, termed NC-VQVAE, is given in Figure \ref{fig:NCVQVAE}. We introduce a non-contrastive self-supervised loss to encode semantic information into the discrete latent representation. The encoded semantics are determined by the augmentations used in non-contrastive objective. The intuition is that the SSL loss pushes the representations of original and augmented views closer together, which should structure the discrete latent space in such a way that data with similar characteristics inhabit distinct regions. Additionally, we add a regularizing term by reconstructing augmented views. We hypothesize that this approach enables the model to generalize better to unseen data by allowing the decoder to \textit{see} the augmented views, as well as preventing the encoder from ignoring the augmentations.

\subsection{Stage 1: Tokenization}
\begin{figure}[!ht]
    \includegraphics[scale=0.35]{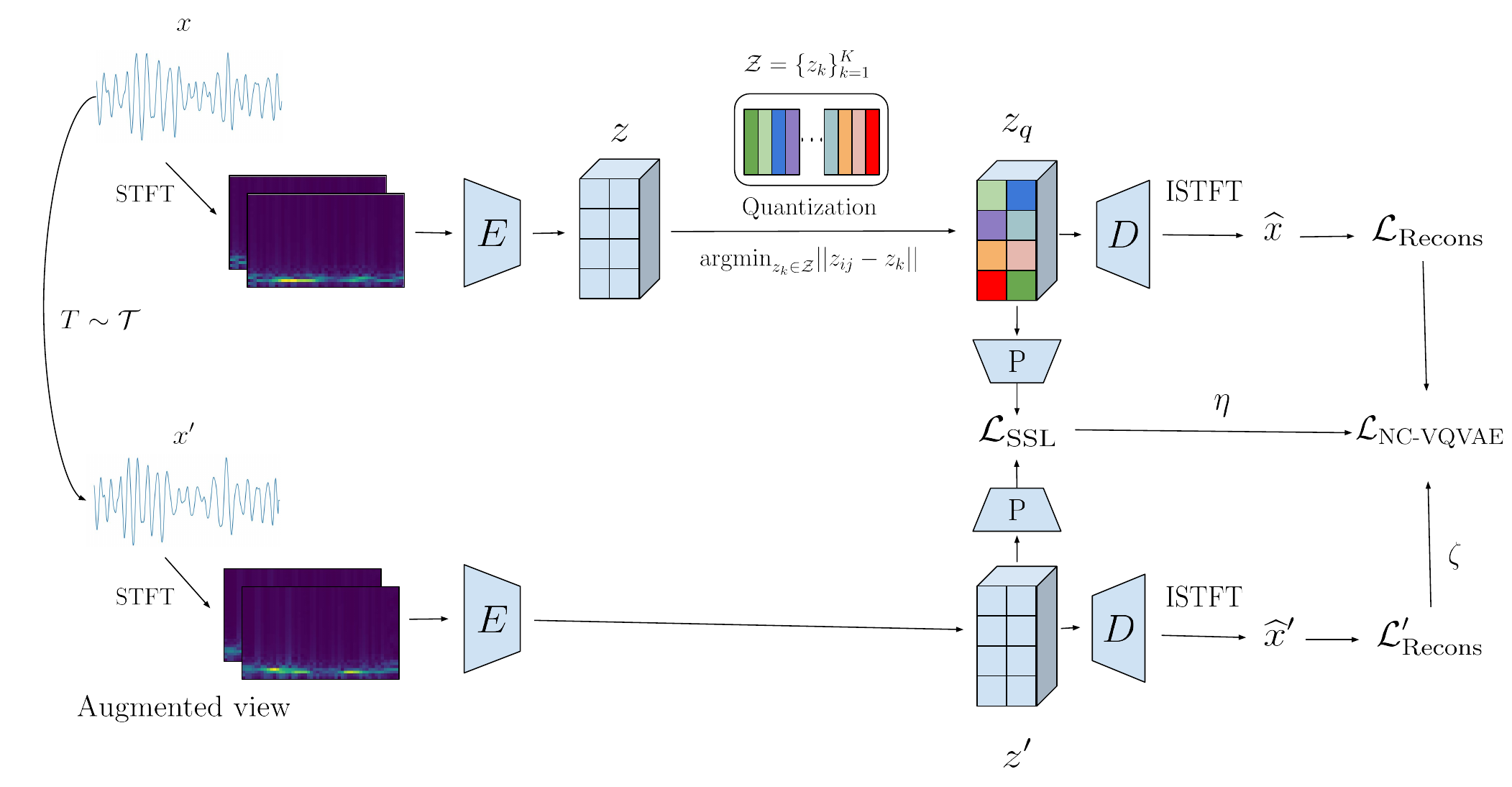}
    \centering  
    \caption{Overview of proposed model: NC-VQVAE.}
    \label{fig:NCVQVAE}
\end{figure}

The architecture of the tokenization model, shown in Figure \ref{fig:NCVQVAE}, consists of two branches: the original and augmented branch. The model takes a time series $x$ as input and creates an augmented view $x'$. The original branch follows the naive VQVAE architecture from \cite{lee2023vector}, while the augmented branch is an autoencoder, constructed by omitting the quantization layer. The views are passed through their respective branches, and we compute the SSL loss derived from the original discrete latent representation $z_q$ and the augmented continuous latent $z'$, before the decoder reconstructs each latent representation. 

The training objective of NC-VQVAE builds on that of TimeVQVAE, without the frequency split. In our setup, the codebook loss is defined as
\begin{equation}
    \label{eq:NCVQVAE_codebook}
    \begin{aligned}
        \loss_\text{codebook} &= ||\sg[z] - z_q ||_2^2 \\
                              &+ \beta||z - \sg[z_q]||_2^2,
    \end{aligned}
\end{equation}
and the reconstruction loss is given by
\begin{equation}
    \label{eq:NCVQVAE_recons}
        \loss_\text{recons} = ||x - \widehat{x} ||_2^2 + ||u - \widehat{u}||_2^2.
\end{equation}
The overall VQ loss is then
\begin{equation}
    \label{eq:NCVQVAE_loss}
    \loss_\text{VQ} = \loss_\text{codebook} + \loss_\text{recons}.
\end{equation}
Our contribution includes the addition of an SSL loss and a reconstruction loss on the augmented branch. The SSL loss varies depending on the method used. We refer to \cite{lee2024computer} and \cite{zbontar2021barlow} for details regarding the VIbCReg and Barlow Twins losses, respectively. 

The augmented reconstruction loss is simply given as 
\begin{equation}
    \label{eq:NCVQVAE_augrecons}
        \loss_\text{recons}' = ||x' - \widehat{x}' ||_2^2 + ||u' - \widehat{u}'||_2^2.
\end{equation}
This loss ensures that the encoder and decoder reconstruct the augmented view, which, in conjunction with the SSL loss, influences the codebook to encode information regarding the augmentations. It also helps prevent the encoder from ignoring reconstruction in favor of the SSL loss. Initial experiments showed that omitting the augmentation reconstruction led to severe overfitting.

The total loss is given by 
\begin{equation}
    \loss_{\text{NC-VQVAE}} = \loss_{\text{VQ}} + \eta\loss_{\text{SSL}} + \zeta \loss_\text{recons}',
\end{equation}
where $\eta$ and $\zeta$ are hyperparameters influencing the importance of each term in the total training objective. 

\subsection{Stage 2: Prior Learning}
Typically, only the codebook indices $k$ from Stage~1 are utilized in Stage~2 while the codebook embeddings in Stage~2 are randomly initialized and learned during the training. However, the learned embeddings in Stage~1 capture both low-level and high-level semantics, containing rich information. We, therefore, use the learned embeddings in Stage~2 so that the prior model can benefit from the learned semantics in Stage~1, unlike the common practice.
Except for this, our Stage~2 is equivalent to MaskGITs.
\section{Experiments}

\subsection{Evaluation Metrics}
Four types of evaluations are used:
\begin{enumerate}
    \item Classification accuracy with Support Vector Machine (SVM) and K-Nearest Neighbors (KNN) on the frozen representations $z_q$
    \item Inception Score (IS)
    \item Fréchet Inception Distance (FID) score
    \item Visual inspection
\end{enumerate}
For classification accuracy, we employ an SVM with a linear kernel and a KNN with 5 neighbors. The SVM evaluates if the learned representations are linearly class-separable, while the KNN measures clustering with respect to the same classes. For calculating IS and FID, we follow the methodology described in \cite{lee2023vector}.


\subsection{Experimental Setup}
\subsubsection{Encoder and decoder}
The encoder and decoder architectures closely resemble those described in \cite{nadavbh12}, with further adaptations from \cite{lee2023vector}.

\subsubsection{VQ}
The VQVAE implementation is based on \cite{VQrepo}. We use a codebook size of 32 and a dimension of 64, employing an exponential moving average with a decay of 0.9 and a commitment loss weight of $\beta = 1$. The codebook embeddings learned in Stage 1 are used to initialize the codebook embeddings in Stage 2.

\subsubsection{SSL}
We implement the projector following the guidelines in \cite{lee2024computer} for both Barlow Twins and VIbCReg. The projectors for both methods are identical, consisting of three linear layers, each with 4096 output units. The first two layers are normalized using \texttt{BatchNorm1d}.

For Barlow Twins the weight of the redundancy reduction term, $\lambda$, is set to $0.005$. The weight of the Barlow Twins loss in NC-VQVAE is set to $\eta = 1/D$ where $D$ is the projector output dimension.

For VIbCReg the weights of the similarity loss, $\lambda$, and variance loss, $\mu$, are both set to 25, while for the covariance loss, $\nu$ is set to 100. The weight of the VIbCReg loss in NC-VQVAE is set to $\eta = 0.01$.

\subsubsection{Prior learning}
Adopting the implementation from \cite{chang2022maskgit}, we set the number of iterations $T$ in the iterative decoding algorithm to 10 and use cosine as mask scheduling function ($\gamma$). The bidirectional transformer is implemented with the following parameters -- a hidden dimension size of 256, 4 layers, 4 attention heads, and a feed-forward ratio of 1.

\subsubsection{Training}
We utilize the AdamW optimizer with batch sizes set to 128 for stage 1 and 256 for stage 2, an initial learning rate of $10^{-3}$, cosine learning rate scheduler, and a weight decay of $10^{-5}$. Both Stage 1 and Stage 2 training procedures run for 1000 epochs.

\subsubsection{Datasets} We evaluate our model on a subset of the UCR Archive presented in Table \ref{tab:UCRsubset}. Each dataset is normalized such that they have zero mean and unit variance.
\begin{table}[!ht]
    \centering
    \caption{The subset of the UCR Archive considered for our experiments.}
    \begin{tabular}{llllll}
    \toprule
    Type      & Name                    & Train & Test & Class & Length \\
    \midrule
    Device    & ElectricDevices         & 8926  & 7711 & 7     & 96     \\
    Sensor    & FordB                   & 3636  & 810  & 2     & 500    \\
    Sensor    & FordA                   & 3601  & 1320 & 2     & 500    \\
    Sensor    & Wafer                   & 1000  & 6164 & 2     & 152    \\
    Simulated & TwoPatterns             & 1000  & 4000 & 4     & 128    \\
    Sensor    & StarLightCurves         & 1000  & 8236 & 3     & 1024   \\
    Motion    & UWaveGestureLibraryAll  & 896   & 3582 & 8     & 945    \\
    ECG       & ECG5000                 & 500   & 4500 & 5     & 140    \\
    Image     & ShapesAll               & 600   & 600  & 60    & 512    \\
    Simulated & Mallat	                & 55	& 2345 & 8	   & 1024   \\
    Image     & Symbols                 & 25    & 995  & 6     & 398    \\
    Sensor    & SonyAIBORobotSurface2   & 27    & 953  & 2     & 65     \\
    Sensor    & SonyAIBORobotSurface1   & 20    & 601  & 2     & 70     \\
    \bottomrule
    \end{tabular}
    \label{tab:UCRsubset}
    \end{table}

\subsubsection{Augmentations} In our experiments we consider three sets of augmentations with different characteristics. 1) \textit{Amplitude Resizing + Window Warp} transforms in both $x$ and $y$ direction. The window warp augmentation randomly selects a section of the time series and speeds it up or down, and the factor is chosen randomly from the interval $[0.9, 2.0]$. We interpolate in order to obtain a time series of equal length as the original. The amplitude resize multiplies the time series by a factor of $1+N(0,0.2)$,
2) \textit{Slice and Shuffle} crops the time series into four randomly selected sections and permutes them, and
3) \textit{Gaussian noise} adds a nose $\epsilon \sim N(0,0.05)$ to each datapoint in the time series.

\section{Results}
We first present results regarding reconstruction and classification capabilities of the learned representations in Stage~1. Afterward, we present IS and FID scores, as well as visual inspection for Stage 2. 
\subsection{Stage 1}
\begin{figure}[!ht]
    \includegraphics[scale=0.3]{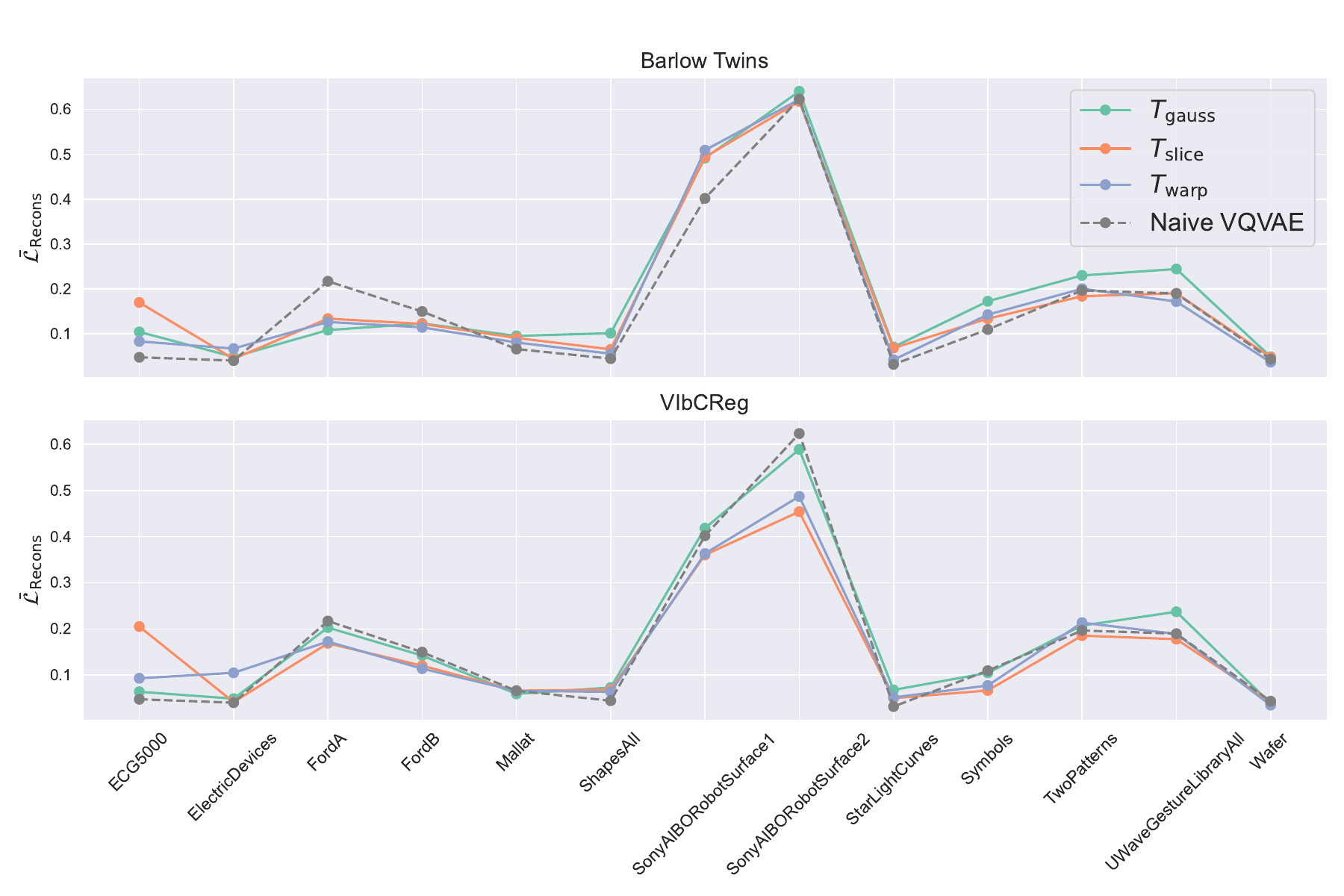}
    \centering  
    \caption{Mean validation reconstruction loss of the models with SSL compared to naive VQVAE.}
    \label{fig:Mean_val_recons}
\end{figure}

Figure~\ref{fig:Mean_val_recons} presents reconstruction loss for the models with SSL and naive VQVAE over the 13 different datasets. We observe that incorporating a non-contrastive loss does not compromise the reconstruction capabilities when compared to the naive VQVAE. 

\begin{figure}[!ht]
    \centering
    
    \begin{subfigure}[b]{\textwidth}
        \centering
        \begin{subfigure}[b]{0.32\textwidth}
            \centering
            \includegraphics[width=\textwidth]{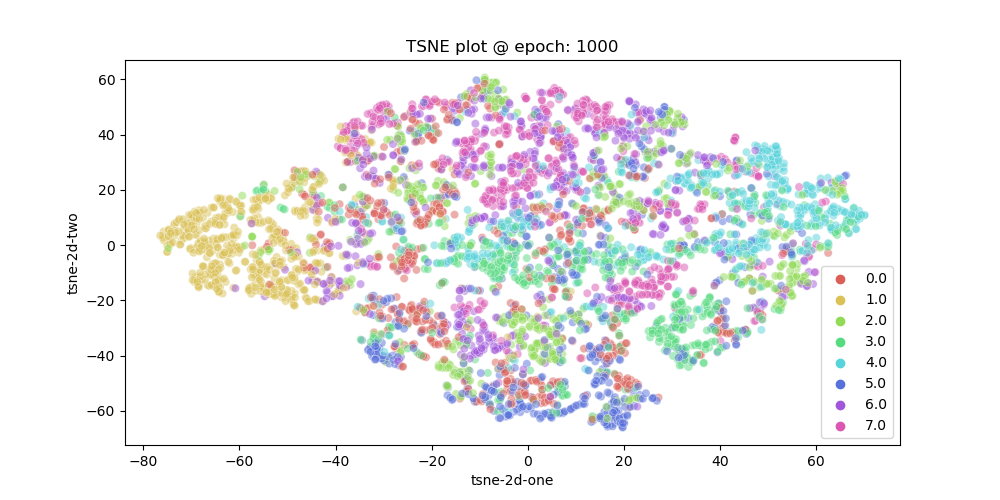}
            \caption{Barlow Twins}
        \end{subfigure}
        \hfill
        \begin{subfigure}[b]{0.32\textwidth}
            \centering
            \includegraphics[width=\textwidth]{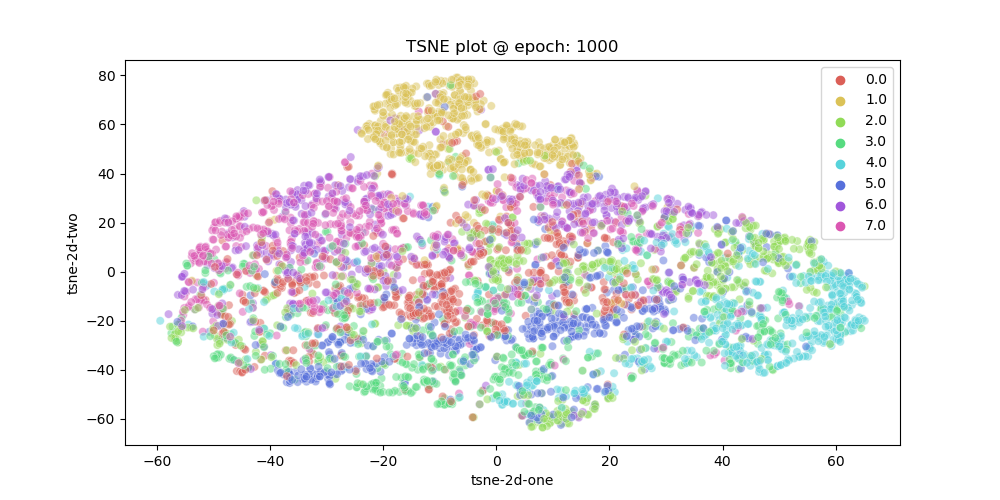}
            \caption{VIbCReg}
        \end{subfigure}
        \hfill
        \begin{subfigure}[b]{0.32\textwidth}
            \centering
            \includegraphics[width=\textwidth]{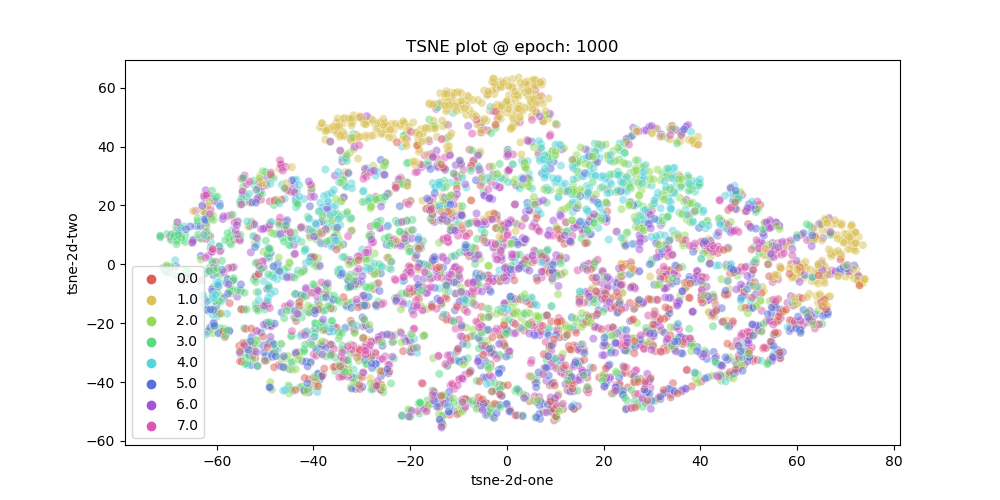}
            \caption{Naive VQVAE}
        \end{subfigure}
        \caption*{UWaveGestureLibraryAll}
        \label{fig:TSNE_UWave}
    \end{subfigure}
    
    \vspace{0.5cm}
    \begin{subfigure}[b]{\textwidth}
        \centering
        \begin{subfigure}[b]{0.32\textwidth}
            \centering
            \includegraphics[width=\textwidth]{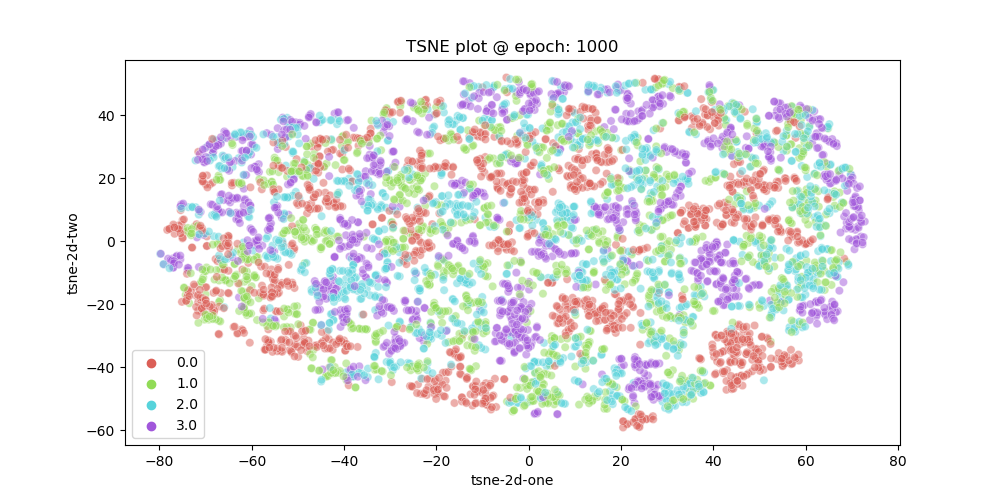}
            \caption{Barlow Twins}
        \end{subfigure}
        \hfill
        \begin{subfigure}[b]{0.32\textwidth}
            \centering
            \includegraphics[width=\textwidth]{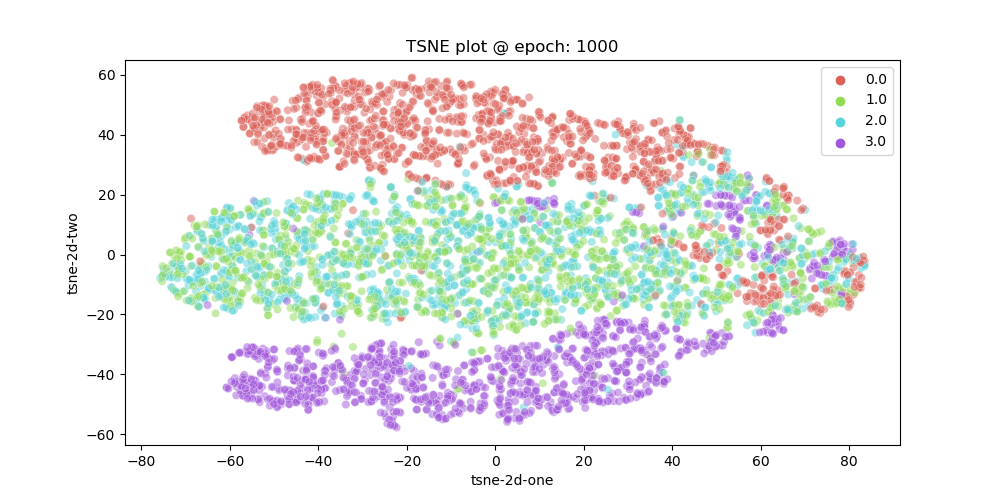}
            \caption{VIbCReg}
        \end{subfigure}
        \hfill
        \begin{subfigure}[b]{0.32\textwidth}
            \centering
            \includegraphics[width=\textwidth]{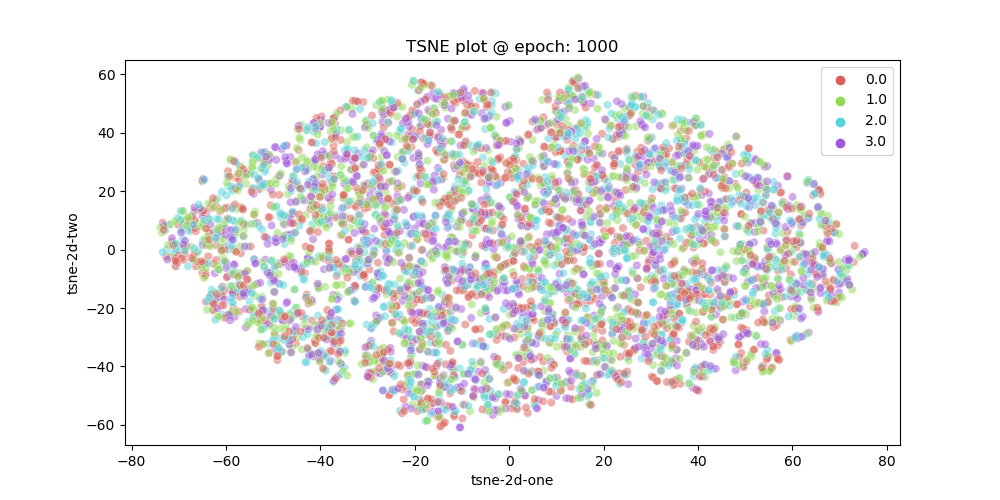}
            \caption{Naive VQVAE}
        \end{subfigure}
        \caption*{TwoPatterns}
        \label{fig:TSNE_TwoPatterns}
    \end{subfigure}
    
    \vspace{0.5cm}
    \begin{subfigure}[b]{\textwidth}
        \centering
        \begin{subfigure}[b]{0.32\textwidth}
            \centering
            \includegraphics[width=\textwidth]{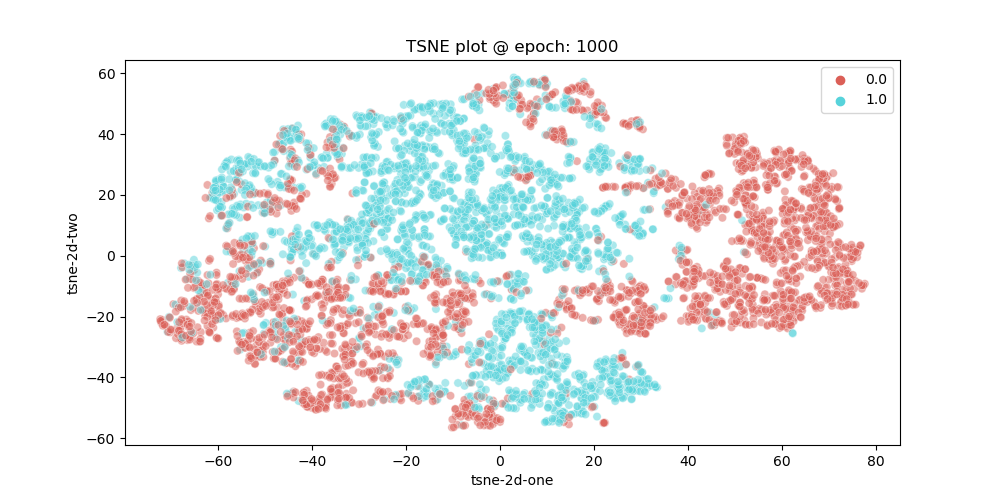}
            \caption{Barlow Twins}
        \end{subfigure}
        \hfill
        \begin{subfigure}[b]{0.32\textwidth}
            \centering
            \includegraphics[width=\textwidth]{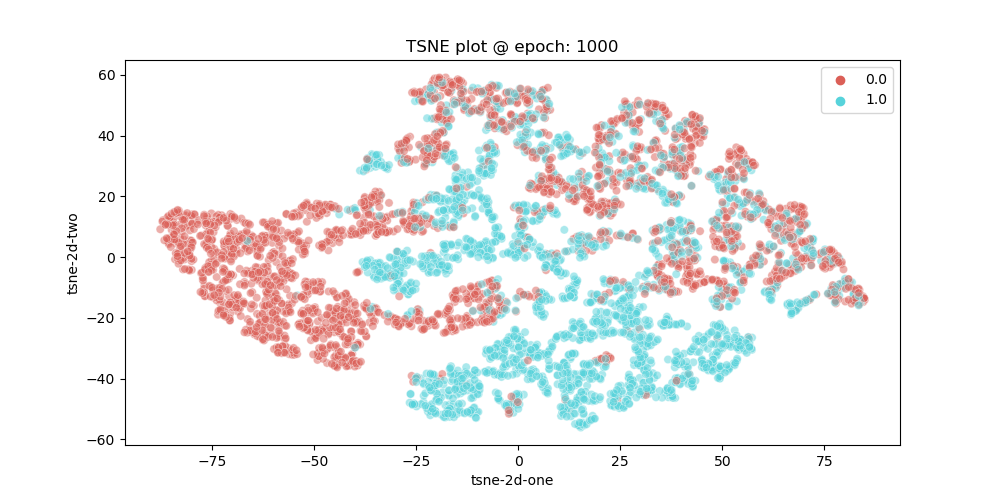}
            \caption{VIbCReg}
        \end{subfigure}
        \hfill
        \begin{subfigure}[b]{0.32\textwidth}
            \centering
            \includegraphics[width=\textwidth]{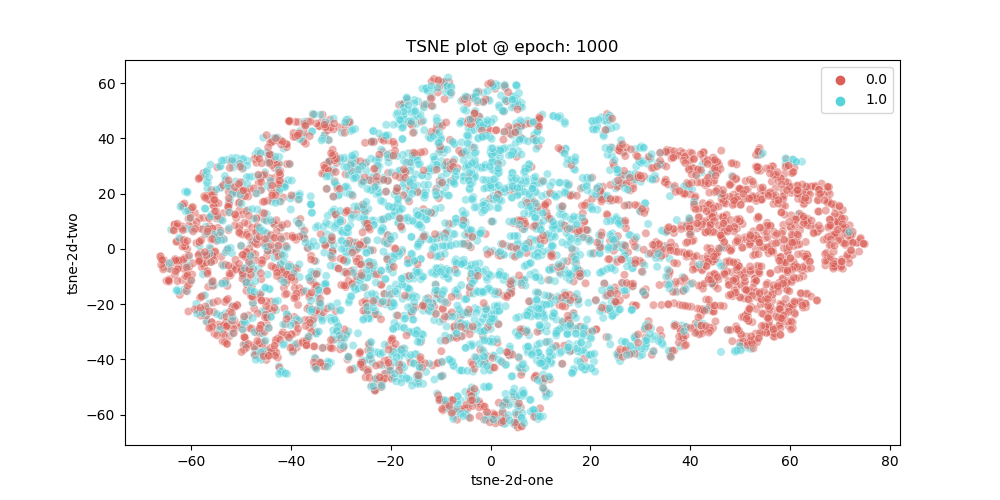}
            \caption{Naive VQVAE}
        \end{subfigure}
        \caption*{FordA}
        \label{fig:FordA_TSNE}
    \end{subfigure}
    
    \caption{t-SNE plot of discrete latent representations from Barlow Twins, VIbCReg, and Naive VQVAE across three datasets: UWaveGestureLibraryAll, TwoPatterns, and FordA. The different colors represent different classes.}
    \label{fig:combined_TSNE}
\end{figure}

Table~\ref{tab:mean_probe} presents the classification accuracy by SVM and KNN. The results clearly show improvement in the accuracy with NC-VQVAE compared to the naive VQVAE. Across 12 out of 13 datasets, a configuration of our model performs best, with the only exception showing a negligible one percent difference for both SVM and KNN. 
This indicates an enhancement in the quality of learned representations in terms of linear class-separability and class-clustering.
The most significant improvements are observed in FordA, FordB, Mallat, ShapesAll, TwoPatterns, and UWaveGestureLibraryAll. 
While the choice of augmentation has a substantial impact, all options lead to significantly improved the accuracy across most datasets. Notably, both SSL methods yield comparable accuracies for a given augmentation, underscoring the importance of selecting appropriate augmentations. 
Further analysis on the augmentations reveals that Slice and Shuffle, as well as Window Warp and Amplitude Resize, result in the most substantial accuracy gains, whereas Gaussian noise consistently yields less pronounced improvements. We hypothesize that since Slice and Warp often generate augmented views that differ considerably from the original, the SSL loss pushes the representations in different directions, potentially leading to better utilization of the latent space. Visualizations in Figure~\ref{fig:combined_TSNE} illustrate the effect of NC-VQVAE on the discrete latent representations of FordA, TwoPatterns and UWaveGestureLibraryAll. These visualizations demonstrate that representations learned with NC-VQVAE exhibit greater structure than those of the naive VQVAE, with similar samples, typically with the same label, clustered closer together in latent space. This suggests that the SSL loss introduces semantic information into the latent representations.

\begin{table}[!t]
    \centering
    \caption{Summary of the mean classification accuracy with KNN and SVM on the learned representations in terms of different SSL methods and augmentations.
    Best scores are highlighted in bold.}
    
    \title{Mean probe accuracy}
    \begin{adjustbox}{width=\textwidth}
    \begin{tabular}{lcc|cc|cc|cc|cc|cc|cc} 
        \toprule
        \multirow{4}{*}{\textbf{Dataset}} & \multicolumn{2}{c}{\textbf{Baseline}} & \multicolumn{12}{c}{\textbf{SSL Method}} \\
                                            \cmidrule(lr){2-3} \cmidrule(lr){4-15}
                                          & \multicolumn{2}{c}{Regular}           & \multicolumn{6}{c}{Barlow Twins}                                                 &  \multicolumn{6}{c}{VIbCReg} \\
                                          \cmidrule(lr){2-3} \cmidrule(lr){4-9} \cmidrule(lr){10-15}
                                          &   \multicolumn{2}{c}{None}            & \multicolumn{2}{c}{Warp}  & \multicolumn{2}{c}{Slice} & \multicolumn{2}{c}{Gauss} & \multicolumn{2}{c}{Warp} & \multicolumn{2}{c}{Slice} & \multicolumn{2}{c}{Gauss} \\
                                          \cmidrule(lr){2-3} \cmidrule(lr){4-5} \cmidrule(lr){6-7} \cmidrule(lr){8-9} \cmidrule(lr){10-11} \cmidrule(lr){12-13}\cmidrule(lr){14-15}
                                          & KNN & SVM                               & KNN & SVM                  & KNN & SVM                & KNN & SVM                 & KNN & SVM                 & KNN & SVM                 & KNN & SVM   \\
        \midrule
        FordA                   & 0.70 & 0.74 & 0.83 & 0.84 & \textbf{0.91} & \textbf{0.89} & 0.80 & 0.83 & 0.80 & 0.74 & 0.87 & 0.86 & 0.76 & 0.78 \\
        ElectricDevices         & 0.35 & 0.41 & 0.35 & \textbf{0.44} & 0.38 & 0.41 & \textbf{0.40} & 0.42 & 0.33 & 0.38 & 0.36 & 0.39 & 0.39 & 0.43 \\
        StarLightCurves         & 0.87 & 0.89 & 0.93 & 0.93 & \textbf{0.94} & \textbf{0.94} & 0.88 & 0.88 & 0.92 & \textbf{0.94} & 0.91 & 0.93 & 0.89 & 0.89 \\
        Wafer                   & 0.93 & 0.89 & 0.96 & \textbf{0.94} & 0.96 & \textbf{0.94} & 0.96 & 0.93 & \textbf{0.97} & 0.94 & 0.96 & 0.92 & \textbf{0.97} & 0.92 \\
        ECG5000                 & 0.80 & 0.83 & 0.85 & 0.81 & \textbf{0.88} & 0.84 & 0.86 & \textbf{0.84} & 0.86 & 0.82 & \textbf{0.88} & \textbf{0.84} & 0.84 & 0.82 \\
        TwoPatterns             & 0.34 & 0.53 & \textbf{0.69} &\textbf{ 0.91} & 0.66 & 0.82 & 0.47 & 0.71 & 0.64 & 0.90 & 0.68 & 0.80 & 0.55 & 0.72 \\
        UWaveGestureLibraryAll  & 0.31 & 0.40 & \textbf{0.62} & 0.70 & 0.56 & 0.63 & 0.40 & 0.54 & \textbf{0.62} & \textbf{0.73} & 0.55 & 0.66 & 0.44 & 0.55 \\
        FordB                   & 0.58 & 0.60 & 0.64 & 0.67 & \textbf{0.74} & \textbf{0.76} & 0.64 & 0.68 & 0.63 & 0.64 & 0.70 & 0.70 & 0.61 & 0.64\\
        ShapesAll               & 0.29 & 0.30 & 0.49 & 0.55 & 0.53 & \textbf{0.60} & 0.40 & 0.48 & 0.48 & 0.56 &\textbf{ 0.54} & \textbf{0.60} & 0.40 & 0.46 \\
        SonyAIBORobotSurface1   & 0.56 & 0.68 & 0.54 & 0.70 & \textbf{0.61} & \textbf{0.74} & 0.53 & 0.70 & 0.48 & \textbf{0.74} & 0.58 & 0.71 & 0.54 & 0.69 \\
        SonyAIBORobotSurface2   & \textbf{0.81} & \textbf{0.86} & 0.77 & 0.79 & 0.80 & 0.80 & 0.80 & 0.81 & 0.77 & 0.85 & 0.80 & 0.85 & 0.80 & 0.85  \\
        Symbols                 & 0.50 & 0.60 & \textbf{0.59} & 0.60 & 0.50 & \textbf{0.66} & \textbf{0.59} & \textbf{0.66} & 0.45 & 0.61 & 0.42 & 0.62 & 0.43 & 0.63 \\
        Mallat                  & 0.63 & 0.77 & 0.72 & 0.81 & 0.76 & 0.83 & 0.68 & 0.78 & \textbf{0.79} & \textbf{0.87} & 0.77 & 0.85 & 0.69 & \textbf{0.86} \\
        \bottomrule
    \end{tabular}
    \end{adjustbox}    
    \label{tab:mean_probe}
\end{table}

\subsection{Stage 2}

Table~\ref{tab:FID_IS_mean} presents the mean scores of IS and FID scores of four runs for the different datasets. The results reveal that our model achieves higher IS scores for 12 out of 13 datasets and lower FID scores for 10 out of 13 datasets.
Upon closer inspection, we observe that VIbCReg demonstrates greater robustness to the choice of augmentation, particularly evident with the Slice and Shuffle augmentation. Furthermore, Gaussian augmentation leads to the most significant improvements across most datasets. The high IS scores suggest that NC-VQVAE produces samples with more accurate class-conditional distributions than the naive VQVAE.
The improved FID scores indicate that the synthetic samples more closely resemble the test data. Additionally, the discrete latent representations from NC-VQVAE provide more class-specific information, as evidenced by the improved downstream classification accuracy observed in Stage~1. This supplementary class-specific information appears to aid the prior learning process in capturing class-conditional distributions more effectively.

\begin{table}[!ht]
    \centering
    \caption{Summary of FID and IS scores in terms of the SSL methods and augmentations. Best mean FID and IS are highlighted in bold.}
    
    \title{Mean FID and IS}
    \begin{adjustbox}{width=\textwidth}
     \begin{tabular}{lcc|cc|cc|cc|cc|cc|cc} 
        \toprule
        \multirow{4}{*}{\textbf{Dataset}} & \multicolumn{2}{c}{\textbf{Baseline}} & \multicolumn{12}{c}{\textbf{SSL Method}} \\
                                            \cmidrule(lr){2-3} \cmidrule(lr){4-15}
                                          & \multicolumn{2}{c}{Regular}           & \multicolumn{6}{c}{Barlow Twins}                                                 &  \multicolumn{6}{c}{VIbCReg} \\
                                          \cmidrule(lr){2-3} \cmidrule(lr){4-9} \cmidrule(lr){10-15}
                                          &   \multicolumn{2}{c}{None}            & \multicolumn{2}{c}{Warp}  & \multicolumn{2}{c}{Slice} & \multicolumn{2}{c}{Gauss} & \multicolumn{2}{c}{Warp} & \multicolumn{2}{c}{Slice} & \multicolumn{2}{c}{Gauss} \\
                                          \cmidrule(lr){2-3} \cmidrule(lr){4-5} \cmidrule(lr){6-7} \cmidrule(lr){8-9} \cmidrule(lr){10-11} \cmidrule(lr){12-13}\cmidrule(lr){14-15}
                                          & FID$\downarrow$ & IS$\uparrow $                             & FID$\downarrow$ & IS$\uparrow$                  & FID$\downarrow$ & IS$\uparrow$                & FID$\downarrow$ & IS$\uparrow$                 & FID$\downarrow$ & IS$\uparrow$                 & FID$\downarrow$ & IS $\uparrow$                 & FID$\downarrow$ & IS$\uparrow$   \\
        \midrule
        FordA                   & 5.15 & 1.16 & 2.59 & 1.41 & 2.36 &\textbf{ 1.45} & \textbf{2.28} & \textbf{1.45} & 3.01 & 1.34 & 2.90 & 1.41 & 3.73 & 1.29 \\
        ElectricDevices         & 13.48 & 3.75 & 16.51 & 3.95 & \textbf{10.20} & 3.93 & 11.54 & 3.75 & 13.99 & \textbf{4.17} & 11.82 & 3.85 & 15.20 & 3.55 \\
        StarLightCurves         & \textbf{1.01} & 1.93 & 1.29 & 2.35 & 1.91 & 2.32 & 1.08 & 2.25 & 1.07 & 2.35 & 1.19 & \textbf{2.36} & 1.05 & 2.22 \\
        Wafer                   & 5.72 & \textbf{1.33} & 3.70 & 1.25 & 4.20 & 1.24 & 2.85 & 1.31 & 3.67 & 1.26 & 3.86 & 1.26 & \textbf{2.84} & 1.31 \\
        ECG5000                 & \textbf{1.62} & 1.94 & 2.61 & \textbf{2.00} & 2.56 & 1.98 & 2.47 & \textbf{2.00} & 2.60 & 1.99 & 2.39 & \textbf{2.00} & 1.76 & 1.99 \\
        TwoPatterns             & 4.04 & 2.41 & 4.00 & 2.54 & 2.96 & 2.66 & \textbf{2.44} & \textbf{2.67} & 4.05 & 2.56 & 3.15 & 2.66 & 2.62 & \textbf{2.67} \\
        UWaveGestureLibraryAll  & 8.48 & 2.13 & 6.77 & 2.86 & 6.64 & 2.96 & 7.35 & 2.73 & 6.80 & 2.91 & \textbf{6.49} & \textbf{2.99} & 7.34 & 2.72 \\
        FordB                   & 4.05 & 1.28 & 2.66 & 1.48 & 3.49 & 1.50 & 2.88 & \textbf{1.52} & \textbf{2.49} & 1.48 & 3.07 & 1.51 & 3.04 & 1.31 \\
        ShapesAll               & \textbf{27.64} & 4.22 & 38.22 & 5.07 & 32.54 & \textbf{5.04} & 32.25 & 4.56 & 36.59 & 4.72 & 35.79 & 4.76 & 31.56 & 4.71 \\
        SonyAIBORobotSurface1   & 23.71 & 1.20 & 30.65 & 1.22 & 31.97 & 1.21 & 25.29 & 1.28 & 26.11 & 1.32 & 28.20 & 1.32 & \textbf{18.61} & \textbf{1.44} \\
        SonyAIBORobotSurface2   & 5.42 & 1.62 & 3.35 & 1.77 & 4.41 & 1.74 & \textbf{1.78} & \textbf{1.81} & 4.43 & 1.74 & 3.32 & 1.79 & 2.36 & 1.79 \\
        Symbols                 & 13.62 & 1.99 & 9.78 & 2.92 & 9.78 & 2.67 & 8.61 & 3.14 & 8.84 & 3.20 & 9.74 & 3.03 & \textbf{8.58} & \textbf{3.24} \\
        Mallat                  & 2.09 & 3.01 & 2.54 & 3.29 & 3.68 & 2.94 & 2.12 & 3.53 & 2.11 & 3.18 & 2.40 & 2.96 & \textbf{1.65} & \textbf{3.72} \\
        \bottomrule
    \end{tabular}
    \end{adjustbox}
    \label{tab:FID_IS_mean}
\end{table}

\subsection{Visual Inspection}
\label{section:visual}
In the following section, each figure displays 50 samples generated from each model. For the ground truth, we plot either 50 randomly selected samples or the entire set if the dataset contains fewer than 50 samples. For datasets with complex distributions or many classes, visually assessing the unconditional distribution is challenging. Therefore, we primarily provide class-conditional samples.
To our surprise, we only observe minor differences in the generated samples from NC-VQVAE trained with different augmentations.

\begin{figure}[!ht]
    \centering
    \begin{minipage}[b]{0.32\textwidth}
        \centering
        \includegraphics[width=\textwidth]{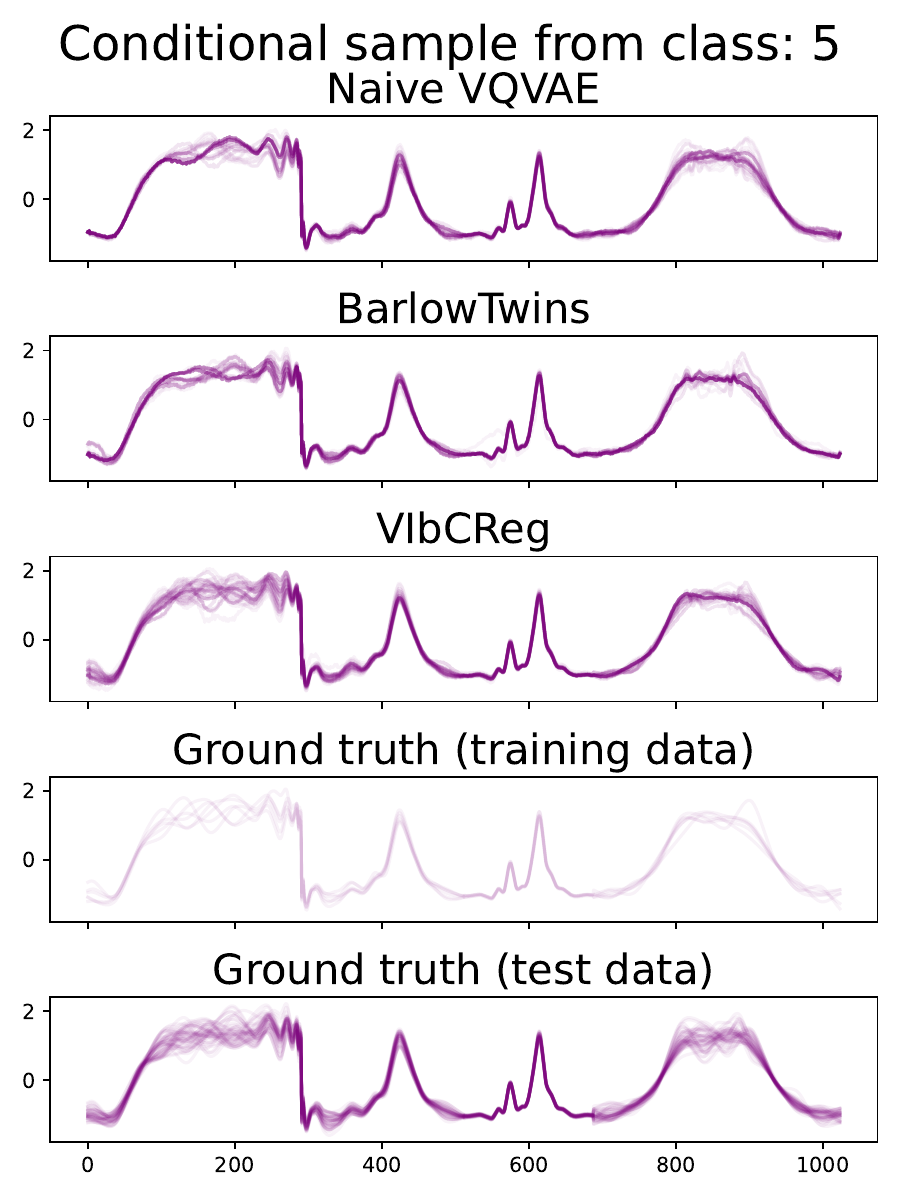}
    \end{minipage}
    \begin{minipage}[b]{0.32\textwidth}
        \centering
        \includegraphics[width=\textwidth]{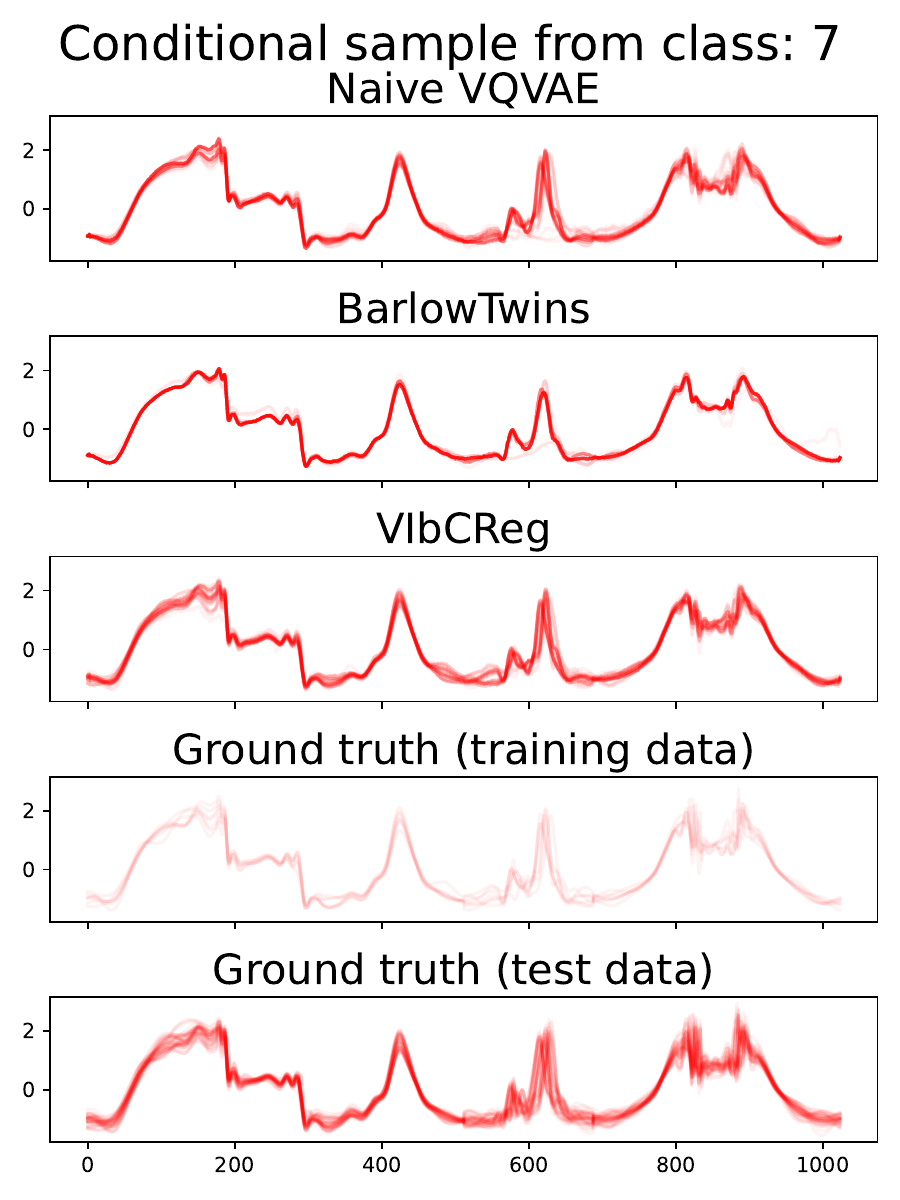}
    \end{minipage}
    \begin{minipage}[b]{0.32\textwidth}
        \centering
        \includegraphics[width=\textwidth]{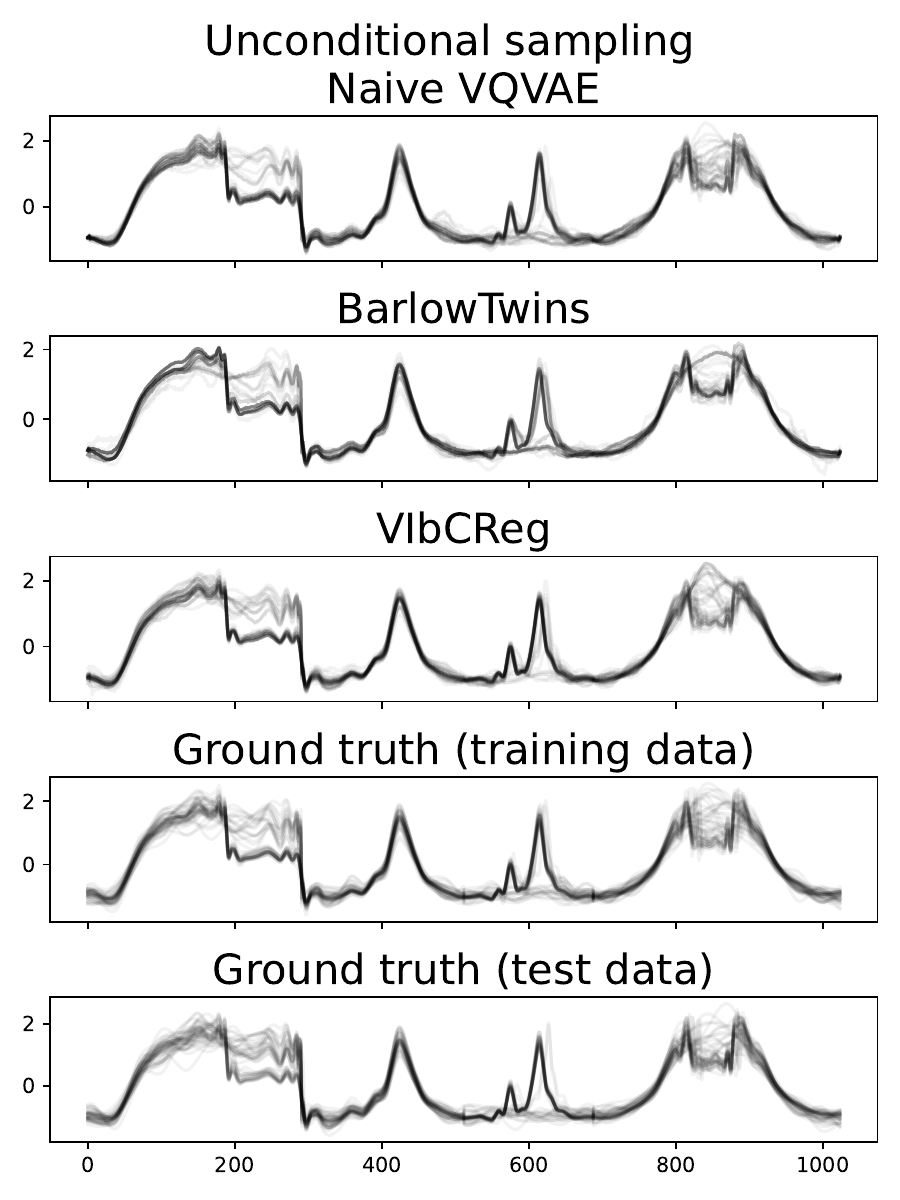}
    \end{minipage}
    \caption{Class conditional distribution for some selected classes of Mallat, in addition to unconditional samples. Barlow and VIbCReg both trained with Gaussian augmentation.}
    \label{fig:Gaussian_Mallat}
\end{figure}

\begin{figure}[!ht]
    \centering
    \begin{minipage}[b]{0.30\textwidth}
        \centering
        \includegraphics[width=\textwidth]{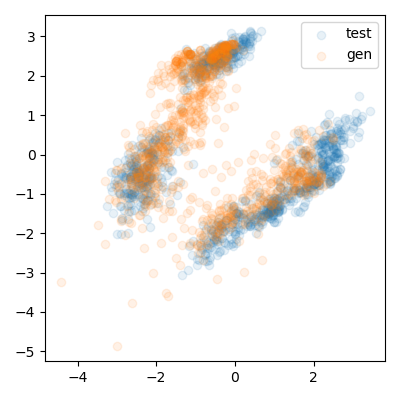}
        \caption*{VIbCReg}
    \end{minipage}
    \begin{minipage}[b]{0.30\textwidth}
        \centering
        \includegraphics[width=\textwidth]{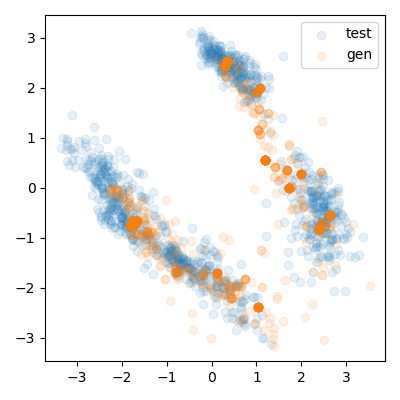}
        \caption*{Barlow Twins}
    \end{minipage}
        \begin{minipage}[b]{0.30\textwidth}
        \centering
        \includegraphics[width=\textwidth]{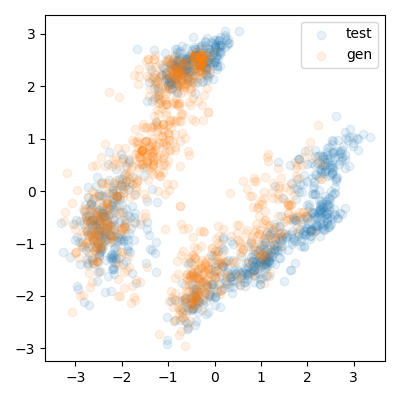}
        \caption*{Naive VQVAE}
    \end{minipage}
    \caption{PCA of discrete latent representation from Mallat. Both VIbCReg and Barlow Twins are trained with Gaussian augmentation.}
    \label{fig:Mallat_latent_PCA}
\end{figure}

\paragraph{Mallat} is a simulated dataset, where the classes have very little variability and training and test distribution are almost indistinguishable, except for sample size.
We observe that VIbCReg is superior in capturing the variability, compared to Barlow Twins and naive VQVAE. This is evident in the first 300 timesteps of class 5 in Figure \ref{fig:Gaussian_Mallat}. Looking as class 7, we see Barlow Twins completely collapsing, essentially producing the identical samples. By inspecting the PCA plots of the discrete latent representations of samples from Mallat, compared to synthetic samples in Figure~\ref{fig:Mallat_latent_PCA}, we see a clear case of representation collapse for Barlow Twins. We hypothesize that the variance term in VIbCReg assists in maintaining variability in the representations, making it more effective in avoiding this type of collapse. 

\begin{figure}[!ht]
    \centering
    \begin{minipage}[b]{0.32\textwidth}
        \centering
        \includegraphics[width=\textwidth]{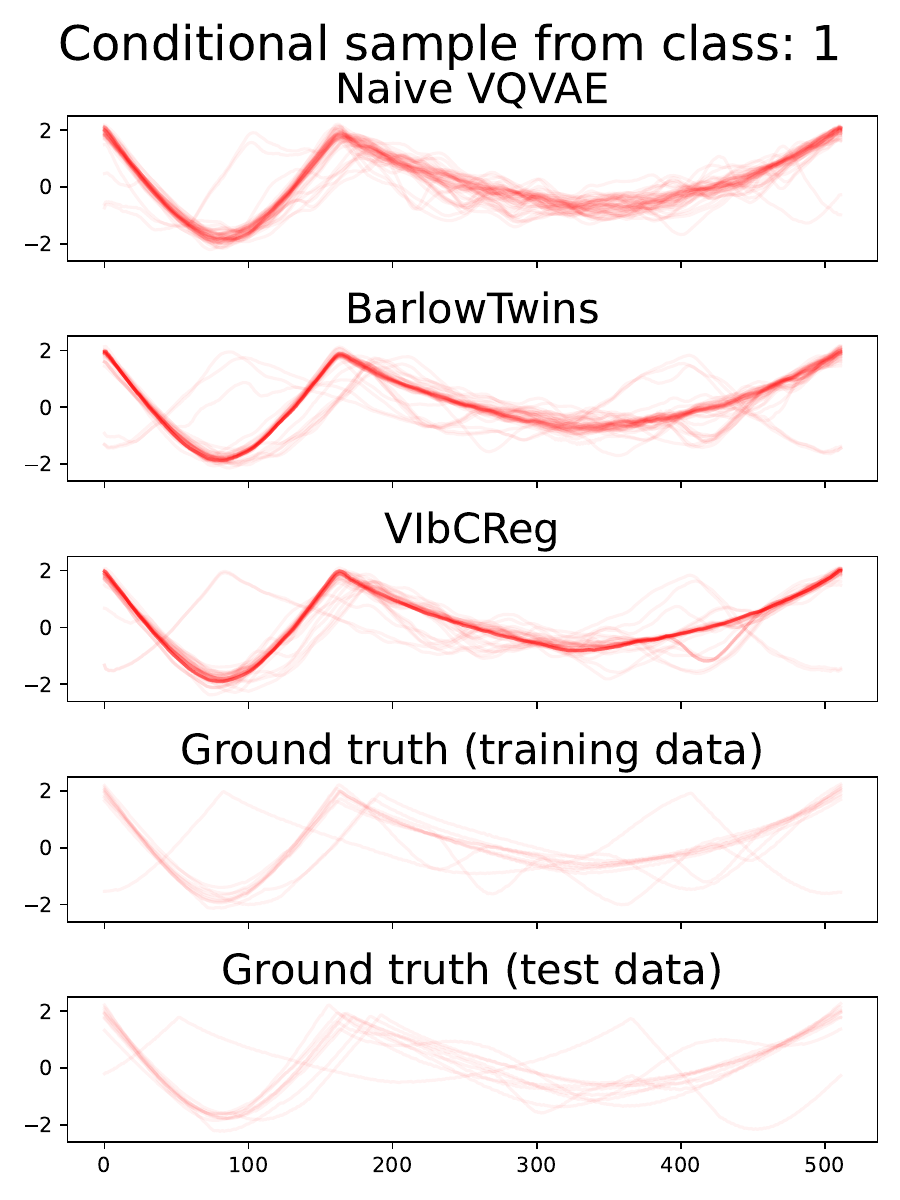}
    \end{minipage}
    \begin{minipage}[b]{0.32\textwidth}
        \centering
        \includegraphics[width=\textwidth]{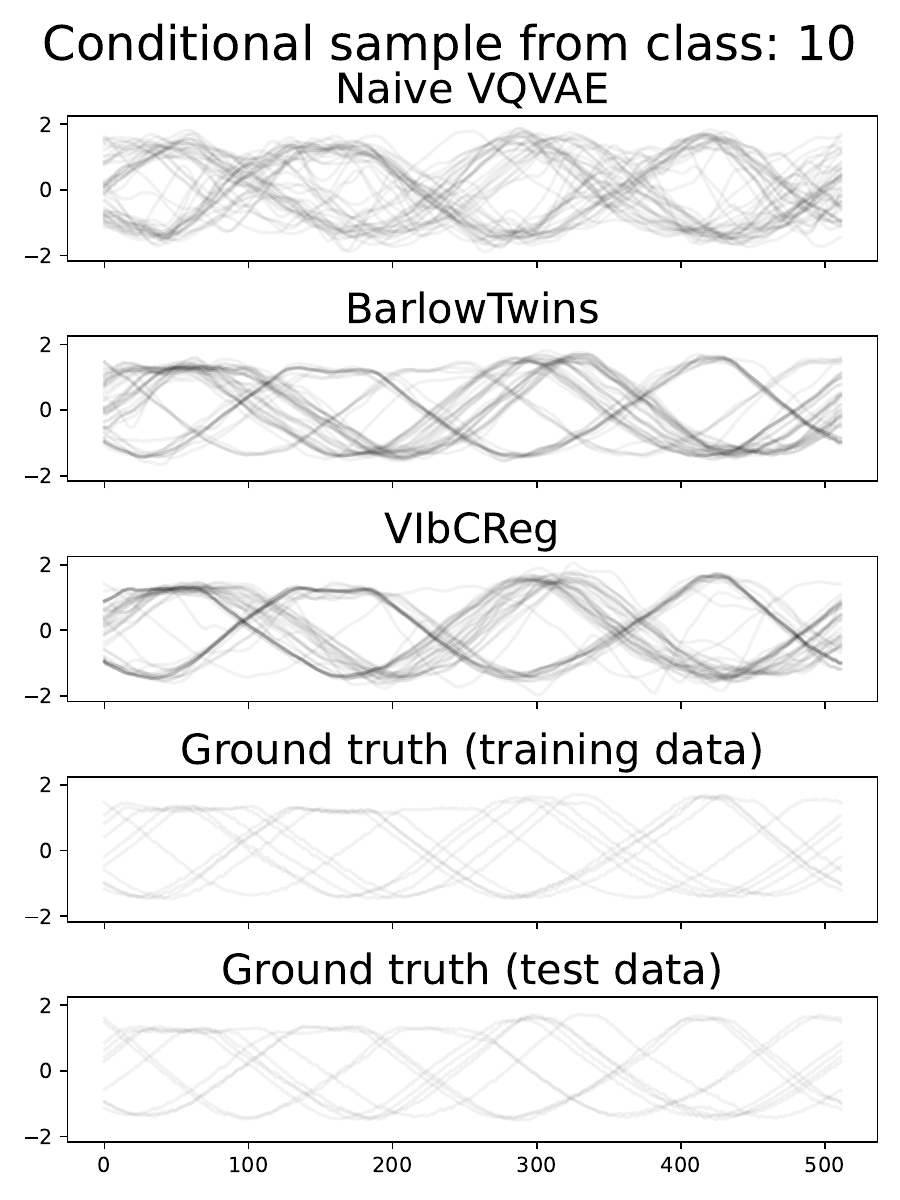}
    \end{minipage}
    \begin{minipage}[b]{0.32\textwidth}
        \centering
        \includegraphics[width=\textwidth]{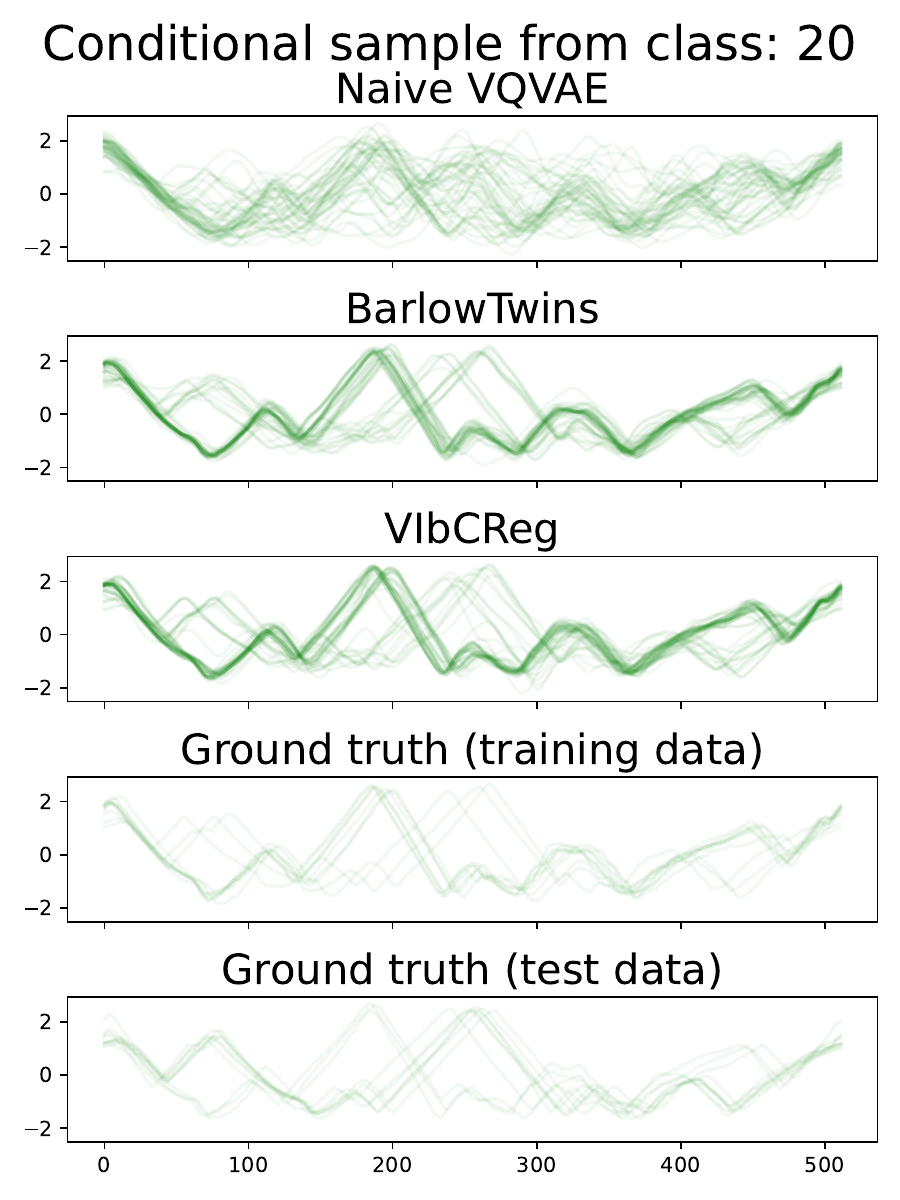}
    \end{minipage}
    \caption{Class conditional distribution for selected classes of ShapesAll. Barlow Twins and VIbCReg are both trained with Gaussian augmentation.}
    \label{fig:Gauss_ShapesAll}
\end{figure}

\paragraph{ShapesALL} consists of 60 classes, with 10 samples within each class. Each class has distinct patterns, with varying complexity. In Figure~\ref{fig:Gauss_ShapesAll}, we clearly observe that naive VQVAE struggles with capturing the global consistency of the samples. Both Barlow Twins and VIbCReg improve IS, but fail to improve FID. By inspecting the samples, it is not evident why NC-VQVAE fails to improve FID. As FID is calculated from unconditional samples, the issue is likely due to an issue not observable from the conditional samples. 

\begin{figure}[!ht]
    \centering
    \begin{minipage}[b]{0.32\textwidth}
        \centering
        \includegraphics[width=\textwidth]{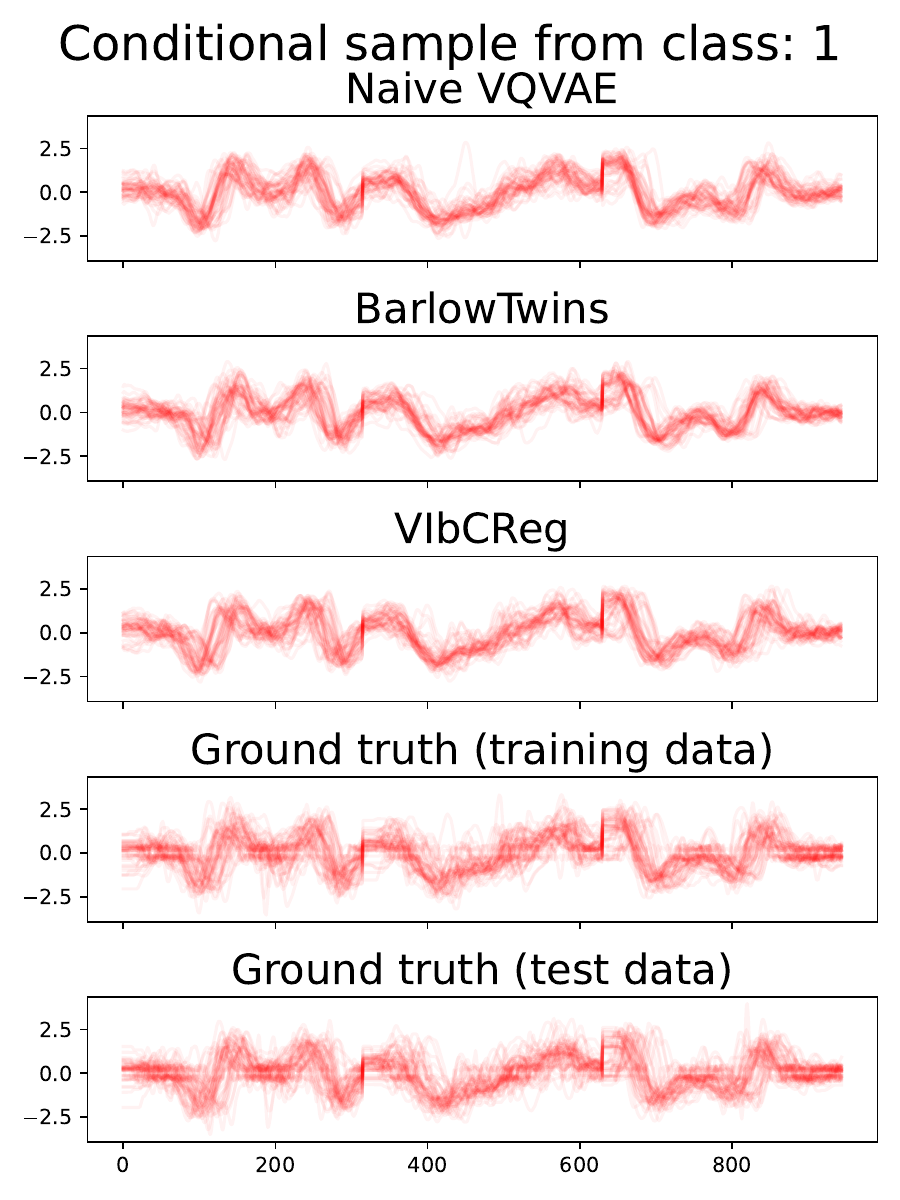}
    \end{minipage}
    \begin{minipage}[b]{0.32\textwidth}
        \centering
        \includegraphics[width=\textwidth]{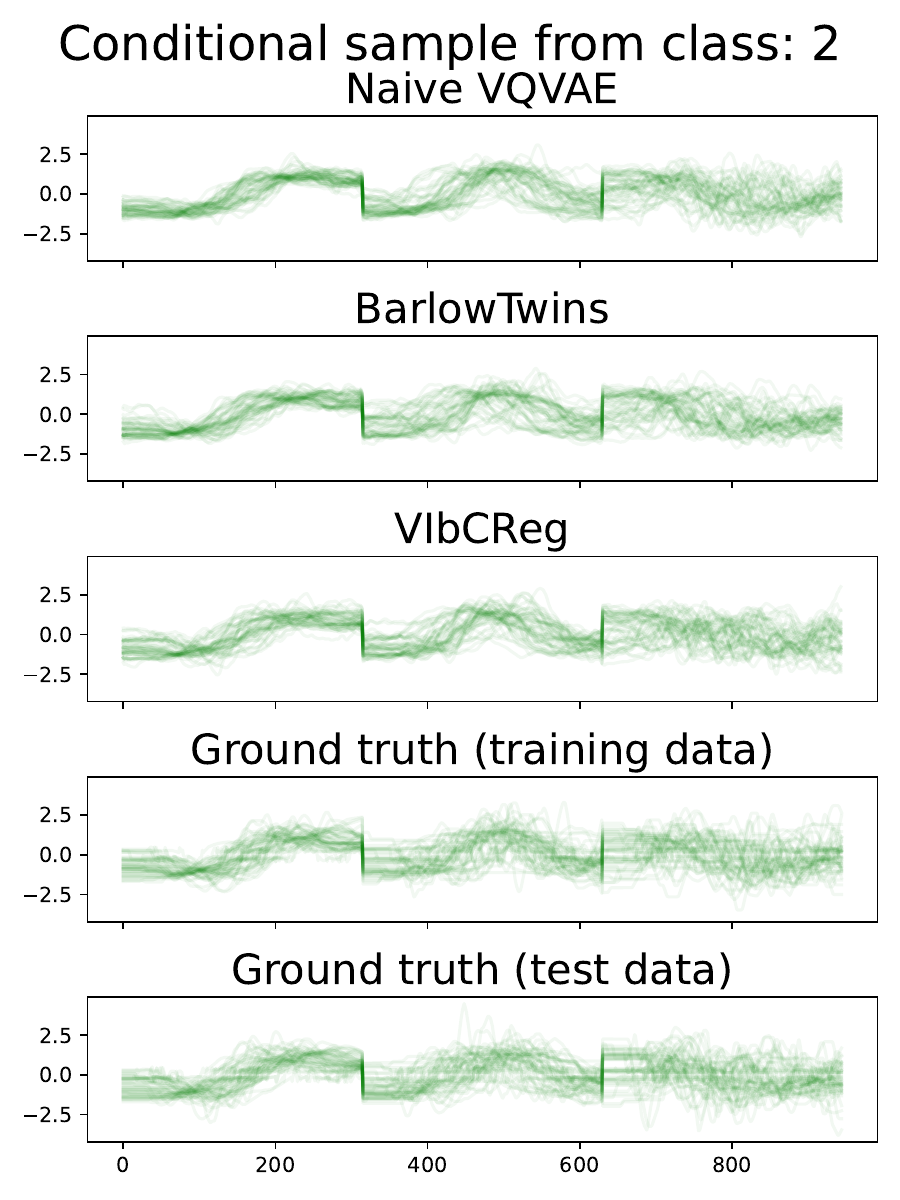}
    \end{minipage}
    \begin{minipage}[b]{0.32\textwidth}
        \centering
        \includegraphics[width=\textwidth]{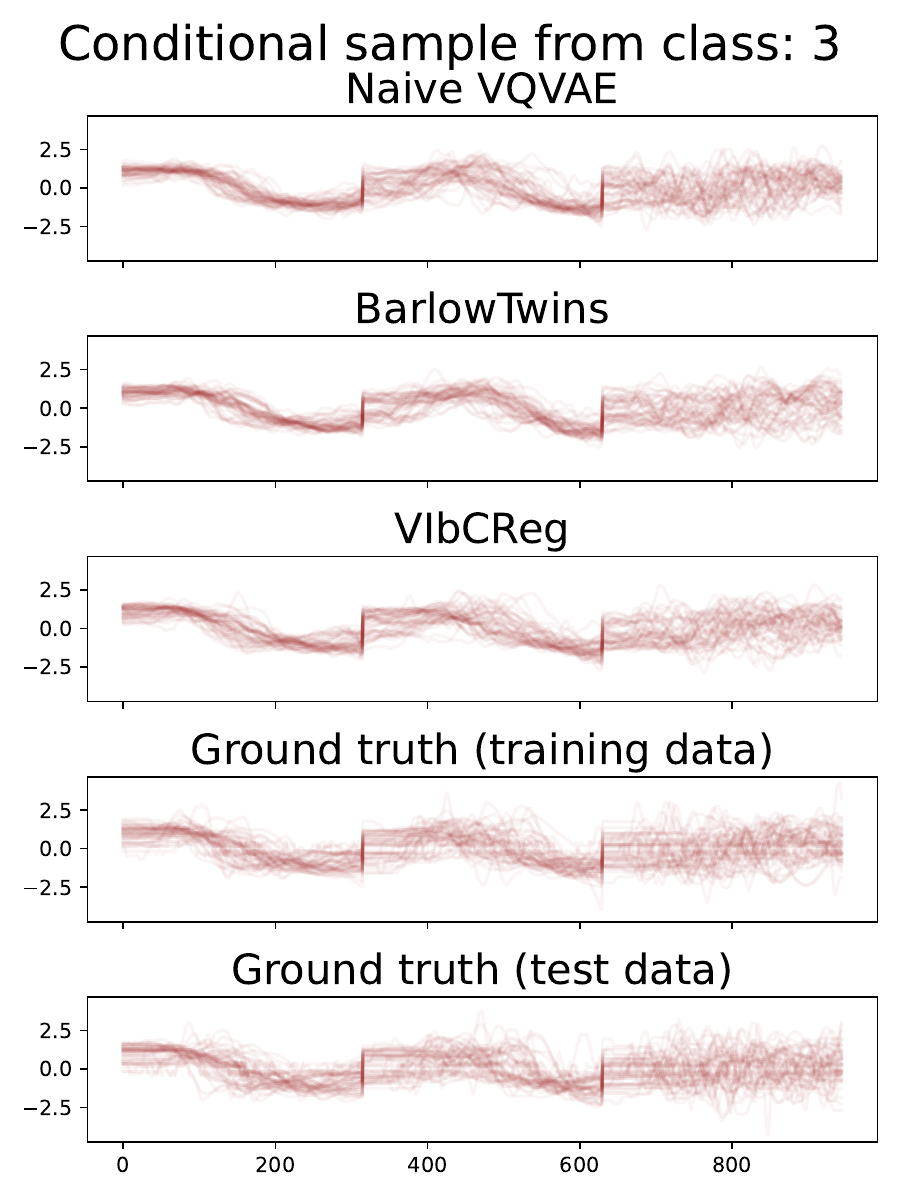}
    \end{minipage}
    \caption{Class conditional distribution for selected classes of UWaveGestureLibraryAll. Barlow and VIbCReg are both trained with Window Warp and Amplitude Resize augmentations.}
    \label{fig:Warp_Uwave}
\end{figure}

\begin{figure}[!ht]
    \centering
    \begin{minipage}[b]{0.29\textwidth}
        \includegraphics[width=\textwidth]{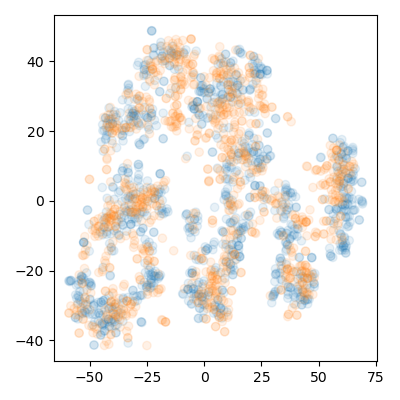}
        \caption*{VIbCReg}
    \end{minipage}
    \begin{minipage}[b]{0.29\textwidth}
        \includegraphics[width=\textwidth]{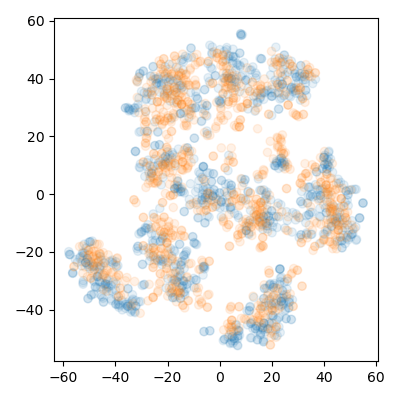}
        \caption*{Barlow Twins}
        \label{fig:TSNE_BT_UWave}
    \end{minipage}
    \begin{minipage}[b]{0.29\textwidth}
        \includegraphics[width=\textwidth]{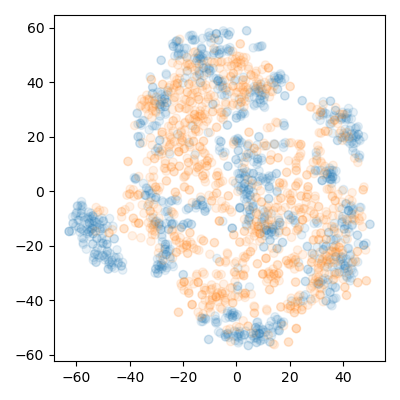}
        \caption*{Naive VQVAE}
        \label{fig:TSNE_Naive_UWave}
    \end{minipage}
    \caption{t-SNE of generated (orange) and test data (blue) on the dataset of UWaveGestureLibraryAll.}
    \label{fig:TSNE_Warp_Uwave}
\end{figure}

\paragraph{UWaveGestureLibraryAll} contains time series with distinct discontinuities and sharp changes in modularity. As noted in \cite{lee2023vector}, such datasets are challenging to model.
In Figure~\ref{fig:Warp_Uwave}, a selected subset of classes is illustrated. We observe upon close inspection that VIbCReg maintains variability in the samples to a greater degree than Barlow Twins, as well as slightly better capturing the dead spots following the discontinuities. 
By investigating the t-SNE plots in Figure~\ref{fig:TSNE_Warp_Uwave}, it becomes more evident that NC-VQVAE captures the distribution better than naive VQVAE.

\section{Conclusion}

The results indicate that NC-VQVAE has clear advantages over the naive VQVAE. Our model can achieve high-quality reconstruction, significantly improve downstream classification accuracy, and can efficiently perform clustering. NC-VQVAE better captures conditional distributions, as evidenced by higher IS scores, and produces higher-quality samples compared to ground truth data, as shown by lower FID scores. Visual inspections further demonstrate that NC-VQVAE achieves better mode coverage and captures global consistency more effectively than the naive VQVAE. Despite a moderate decrease in FID, the quality of the generated samples is often much higher. 

We conclude that the expressive representations learned by NC-VQVAE assist in learning class-specific details and high level semantics, thereby enhancing the quality of synthetic samples.

\begin{credits}
\subsubsection{\ackname} We would like to thank the Norwegian Research Council for funding the Machine Learning for Irregular Time Series (ML4ITS) project (312062). This funding directly supported this research. We also would like to thank all the people who have contributed to the UCR time series classification archive.
\end{credits}
%
%
%
\bibliographystyle{splncs04}
\bibliography{main}

\end{document}